\def\year{2023}\relax
\documentclass[letterpaper]{article} 
\usepackage{aaai23}  
\usepackage{times}  
\usepackage{helvet}  
\usepackage{courier}  
\usepackage[hyphens]{url}  
\usepackage{graphicx} 
\usepackage{natbib}  
\usepackage{caption} 
\DeclareCaptionStyle{ruled}{labelfont=normalfont,labelsep=colon,strut=off} 
\usepackage{algorithm}
\usepackage{algpseudocode}

\usepackage{booktabs}
\usepackage{amsfonts}       
\usepackage{nicefrac}       
\usepackage{microtype}      
\usepackage{xcolor}         

\usepackage{amsmath}       
\usepackage{mathtools}
\usepackage{amsthm}
\usepackage{amssymb}
\usepackage{ifthen}
\usepackage{bm}

\DeclareMathOperator*{\argmax}{arg\,max}

\usepackage{newfloat}
\usepackage{listings}
\floatstyle{ruled}
\newfloat{listing}{tb}{lst}{}
\floatname{listing}{Listing}
\nocopyright
\title{Exploration via Epistemic Value Estimation}
\author {
    Simon Schmitt,\textsuperscript{\rm 1,2}
    John Shawe-Taylor, \textsuperscript{\rm 2}
    Hado van Hasselt\textsuperscript{\rm 1}
}
\affiliations {
    \textsuperscript{\rm 1}DeepMind \\
    \textsuperscript{\rm 2}University College London, UK \\
    suschmitt@google.com
}

\newcommand{\defeq}{\vcentcolon=}
\newcommand{\trans}[1]{{#1}^\top}
\newcommand{\Expectation}[2][]{\mathbb{E}_{#1}\left[#2\right]}
\newcommand{\expectation}[2][]{\mathbb{E}_{#1}[#2]}
\newcommand{\Variance}[2][]{\mathbb{V}_{#1}\left[#2\right]}
\newcommand{\Fisher}[2][]{\mathcal{I}_{#1}(#2)}
\newcommand{\EFisher}[1]{\mathcal{\hat{I}}(#1)}
\newcommand{\Hessian}[1]{\mathcal{H}(#1)}

\newcommand{\gradtheta}{\nabla_\theta}

\newcommand{\Data}{\mathcal{D}}
\newcommand{\trajectory}{\tau}

\newcommand{\return}{G}
\newcommand{\returnn}{\return^\bootstrapn}
\newcommand{\mcreturn}{Z}

\newcommand{\loss}{\mathcal{L}}

\newcommand{\qtheta}{q_\theta}
\newcommand{\qthetasa}{\qtheta(s, a)}
\newcommand{\sample}[1]{{#1'}}
\newcommand{\thetasample}{\sample{\theta}}
\newcommand{\thetamle}{\theta^{\mathrm{MLE}}}
\newcommand{\thetamlen}{\theta^{\mathrm{MLE}}_n}
\newcommand{\thetamap}{\theta^{\mathrm{MAP}}}
\newcommand{\thetamapn}{\theta^{\mathrm{MAP}}_n}
\newcommand{\thetatrue}{\theta^\mathrm{True}}
\newcommand{\thetatarget}{\bar{\theta}}
\newcommand{\normal}[1]{\mathcal{N}(#1)}
\newcommand{\Normal}[1]{\mathcal{N}\left(#1\right)}
\newcommand{\modelf}[1]{f_{#1}}
\newcommand{\sigmareturn}{\sigma^{\mathrm{Return}}}

\newcommand{\grad}[1]{g^{\mathrm{#1}}}

\newcommand{\gradmle}{\grad{MLE}}
\newcommand{\gradll}{\grad{LogL}}
\newcommand{\gradllt}[1]{\gradll_t(#1)}
\newcommand{\Fdiag}{f^{\mathrm{diagonal}}}
\newcommand{\Funbiased}{f^{\mathrm{unbiased}}}
\newcommand{\explorationscale}{\omega}
\newcommand{\bootstrapn}{k}

\newcommand{\mcreturnseq}{\bm{\mcreturn^n}}

\newcommand{\stateseq}{\bm{S^n}}

\newcommand{\mattovec}{\mathrm{vec}}
\newcommand{\vectomat}{\mathrm{mat}}

\begin{document}

\maketitle

\begin{abstract}
How to efficiently explore in reinforcement learning is an open problem. Many exploration algorithms employ the epistemic uncertainty of their own value predictions -- for instance to compute an exploration bonus or upper confidence bound. Unfortunately the required uncertainty is difficult to estimate in general with function approximation.

We propose epistemic value estimation (EVE): a recipe that is compatible with sequential decision making and with neural network function approximators. It equips agents with a tractable posterior over all their parameters from which epistemic value uncertainty can be computed efficiently.

We use the recipe to derive an epistemic Q-Learning agent and observe competitive performance on a series of benchmarks. Experiments confirm that the EVE recipe facilitates efficient exploration in hard exploration tasks.

\end{abstract}

\section{Introduction}
Reinforcement learning agents strive to maximize return in sequential decision making problems.
To learn about the problem structure they take potentially costly exploratory actions. Ideally the price of exploration will be outweighed by future gains afforded by the gained information. Balancing those two conflicting objectives is called the \emph{exploration versus exploitation trade-off}. It is at the heart of reinforcement research and has been studied extensively~\cite{Bellman:1957, Martin:1967,Duff:2002,Guez2012bayesadaptive}.
Many proposed exploration algorithms are uncertainty based: To explore efficiently the uncertainty of the value prediction is used -- for instance as an exploration bonus or upper confidence interval \citep{Auer:2002}, or for Thomson Sampling~\citep{thomson1933}.

\paragraph{A Recipe for Uncertainty Estimation}
How to measure value uncertainty in deep reinforcement learning is actively researched and so far no consensus has been reached. We propose \emph{epistemic value estimation} (EVE), a principled and computationally efficient recipe for uncertainty estimation with function approximation. 
It is compatible with neural networks and specifically supports off-policy learning and value-bootstrapping -- key concepts distinguishing reinforcement learning from supervised learning.

EVE equips the agent with a tractable posterior over all its agent parameters, from which epistemic value uncertainty can be computed efficiently. Considering all agent parameters distinguishes it from prior approaches that are Bayesian only on the final network layer.
The recipe reinterprets value prediction as density estimation of returns. Drawing on insights from parametric statistics it then approximates the posterior over agent parameters using a specifically structured Gaussian distribution.
Besides being motivated in statistical theory this approximation is efficient to sample from and convenient to estimate using automatic differentiation frameworks. As a result it obtains favourable computational performance compared to ensemble methods that need to store and update multiple models concurrently.

When applied to Q-Learning, it matches the exploration performance of Bootstrapped DQN on the Deep Sea benchmark -- providing encouraging evidence for our recipe.

\paragraph{What are Epistemic Values?}
As we are uncertain about the true value (i.e. the expected return) we can treat it as a random variable given the agents experience. We will call the corresponding random variable the \emph{epistemic value} to emphasize that the value distribution differs from the return distribution. On average the epistemic value uncertainty decreases with more observed data while the return uncertainty is by definition invariant to it.
$$ 
    \underbrace{p(V | s, \Data)}_{\substack{
        \mathrm{Posterior\ of\ \mathbf{mean}\ returns\ } V \\ \mathrm{captures\ \mathbf{epistemic}\ uncertainty.}
    }} \neq  
    \underbrace{p(\mcreturn| s, \Data)}_{\substack{
        \mathrm{Density\ of\ \mathbf{Monte-Carlo}\ returns \ } Z \\
        \mathrm{captures\ \mathbf{aleatoric}\ uncertainty.} \\
    }}
$$
For example the near-optimal UCB algorithm for bandits upper-bounds the uncertainty of the epistemic value $V$. If it were to instead upper-bound the uncertainty of return $Z$ it would continually keep selecting the action with the noisiest return and cease to be optimal. The EVE recipe provides epistemic value estimates for sequential decision making problems with function approximation.

\begin{figure*}[h]
\centering
\includegraphics[width=0.95\textwidth]{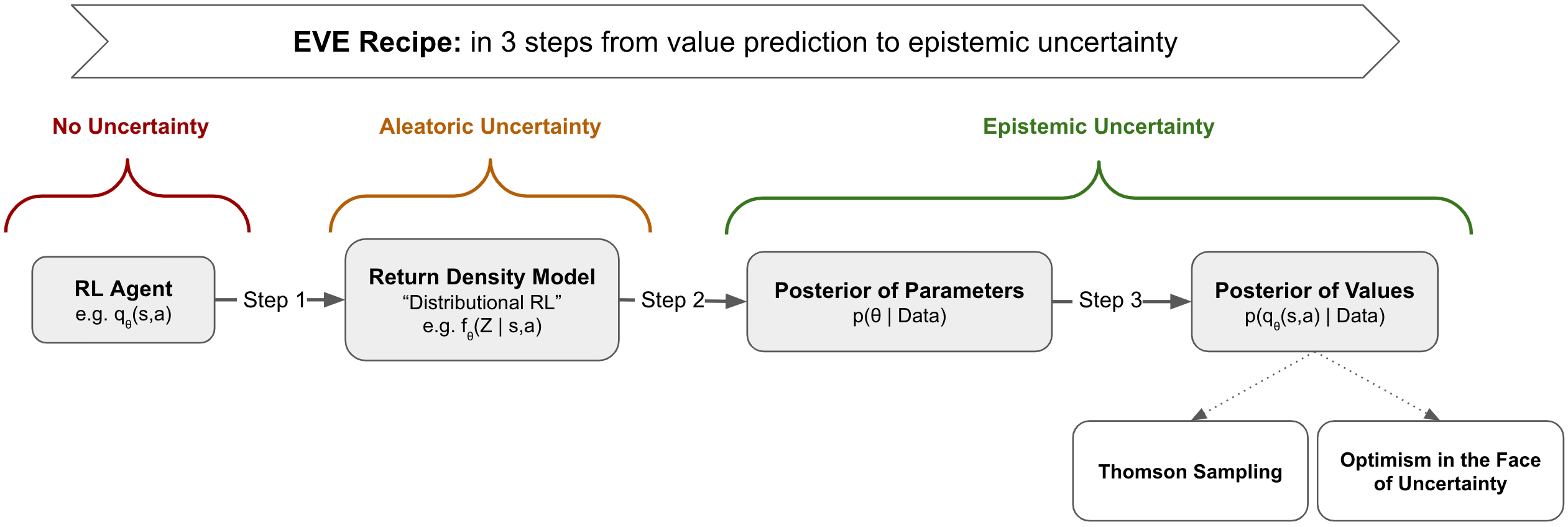}
\caption{
The Epistemic Value Estimation recipe (EVE) equips agents with a tractable posterior over all their parameters from which epistemic uncertainty can be computed efficiently. The first step converts the agent into a distributional agent, that captures the seemingly unrelated aleatoric return uncertainty. The second step approximates the posterior over all agent parameters from which the third step can obtain the epistemic value uncertainty $p(q_\theta(s, a) | \Data)$.
}
\label{fig:recipe_diagram}
\end{figure*}

\section{Background}
We investigate exploration in Markov Decision Processes~\cite{Bellman:1957}. We largely follow the notation from~\citet{Sutton:98book} denoting actions $A_t$ taken at states $S_t$ yielding new states $S_{t+1}$ and rewards $R_{t+1}$. When selecting actions $A_t \sim \pi(S_t)$ from a policy $\pi$ the Monte-Carlo return at state $S_t$ is $\mcreturn_t \defeq \sum_{i=0}^\infty \gamma^i R_{t+i+1}$ to be distinguished from bootstrapped returns $\return_t$.
The expected return is called the value $v_\pi(s) \defeq \Expectation[\pi]{\mcreturn_t \mid S_t = s}$.
Function approximation is often used to represent value estimates in large state spaces -- typically by parametric functions $v_\theta(s)$. Given no prior knowledge about the MDP we strive to find a policy that selects actions to maximize the expected return. 

We intend to estimate the uncertainty of predicted values $v_\theta(s)$ given limited data $\Data$ comprised of $n$ steps. In the process we estimate the distribution of Monte-Carlo returns $\mcreturn$. Recall from the introduction that the uncertainty of returns is different from the uncertainty of values (expected returns).

In general, consider some random variables $X_i$ that are sampled i.i.d. from a distribution $\modelf{\thetatrue}$ parameterized by an unknown parameter $\thetatrue$. We can construct a frequentist estimate of $\thetatrue$ using the maximum likelihood estimator:
$ \thetamlen \defeq \argmax_\theta p(X_1, \dots, X_n|\theta) = \argmax_\theta \prod_i \modelf{\theta}(X_i) \,.$
We can also adopt a Bayesian view, and assume a prior belief about the likelihood of parameters $p(\theta)$ and use Bayes rule to compute the posterior:
$ p(\theta| \Data) \propto p(\Data|\theta) p(\theta) $.
The maximum a posteriori solution is then typically considered the best parameter point-estimate:
$ \thetamapn \defeq \argmax_\theta p(\theta| \Data) $
For notational simplicity we drop the $n$ from $\thetamlen, \thetamapn$.
One can show that $\thetamle \to \thetatrue$ and $\thetamap \to \thetatrue$ when $\modelf{}$ and prior $p(\theta)$ satisfy appropriate regularity conditions \citep[see][for details]{vandervaart1998:asymptotic,nickl2012:statistical}. Finally, we use $x \odot y$ to refer to the Hadamard product of two vectors $x, y$.

\section{A Scalable Recipe for Epistemic Uncertainty in RL}
\label{sec:recipe}
Our goal is to formalize a recipe that can provide deep RL agents with epistemic value uncertainty and posterior sampling capabilities in a way compatible with their large neural networks. Our recipe supports off-policy Q-Learning and $\bootstrapn$-step TD learning where an agent bootstraps on its own predictions -- a requirement that differentiates it from supervised learning. We first state the three major steps in the recipe, and then explain each step in turn in the remainder of this section.

\paragraph{Intuition} For an intuitive summary of the three steps consider Figure~\ref{fig:recipe_diagram}.

\paragraph{Steps}
\begin{enumerate}
    \item Use a suitable parametric density model $\modelf{\theta}$ to predict the distribution of returns: $\modelf{\theta}(\mcreturn|s) \approx p(\mcreturn| s)$. 
    \item Using $\modelf{\theta}$, approximate the corresponding posterior of parameters $p(\theta | \Data)$ with a specifically constructed Gaussian distribution.
    \item Finally, compute $p(v_\theta(s)| \Data)$ from $p(\theta | \Data)$.
\end{enumerate}

\subsection{Step 1: Model \& Log-Likelihood}
This step aims at setting up a parametric agent model that predicts the entire distribution of Monte-Carlo returns $p(\mcreturn|s, a)$. We require its log-likelihood gradient for model fitting and for the posterior approximation in Step 2.

\paragraph{Modelling Monte-Carlo Returns}
In reinforcement learning the Monte-Carlo return is not always available or desirable due to its high variance and on-policy nature. In fact many agents use $\bootstrapn$-step bootstrapped returns $\returnn_t \defeq \sum_{i=1}^{\bootstrapn} \gamma^{i-1} R_{t+i} + \gamma^{\bootstrapn-1} v_\theta(S_{t+\bootstrapn})$.
This is an important difference to supervised learning with fixed-outputs pairs.
For us this poses a challenge as they are differently distributed
$$p(\mcreturn|s) \neq p(\returnn|s)$$
and predicting $\bootstrapn$-step returns would only capture the epistemic uncertainty of the first $\bootstrapn$ steps and disregard the uncertainty of the bootstrap value $v_\theta(S_{t+\bootstrapn})$.
A similar problem occurs with value iteration where returns are off-policy and differ from the observed Monte-Carlo return: $\return^Q_t \defeq R_{t+1} + \gamma \max_a Q(S_{t+1}, a)$.

A solution is to make the agent \emph{distributional}~\citep{Dearden:1998,Bellemare2017distributional}:
We sample hypothetical Monte-Carlo returns $\mcreturn'$ such that $p(\mcreturn') = p(\mcreturn)$ and use them instead of $\returnn$ or $\return^Q$.
Rather than bootstrapping $\returnn_t$ with the value $v_\theta(S_{t+\bootstrapn})$ we bootstrap using a Monte-Carlo return sample from our model:
\begin{align}
    \Tilde{\mcreturn}_{t+\bootstrapn} &\sim f_\theta(\mcreturn | S_{t+\bootstrapn}) 
\end{align}
and define a \emph{modelled return} $\mcreturn'_t$ that preserves the distributions $p(\mcreturn'_t | S_t) \approx  p(\mcreturn_t| S_t)$:
\begin{align}
    \mcreturn'_t &\defeq \sum_{i=1}^\bootstrapn \gamma^{i-1} R_{t+i} + \gamma^\bootstrapn \Tilde{\mcreturn}_{t+\bootstrapn}
\end{align}
Sampling from $\modelf{}$ induces an approximation error, however \citet{Bellemare2017distributional} have observed strong empirical performance and provided a theoretical analysis for learning distributional value functions.

\paragraph{Log-Likelihood Estimation}
Given the density model $\modelf{\theta}$ and modelled or actual Monte-Carlo returns $\mcreturn'$ at state $S$ we can estimate the maximum likelihood solution $\thetamle$. Typically this is achieved through stochastic gradient decent on $\gradtheta \log \modelf{\theta}(\mcreturn' | S)$. The same gradient will also be used in Step 2 to approximate the posterior. 

\subsection{Step 2: Posterior Approximation}
\label{sec:recipe_step2_posterior_approximation}

We will now derive and approximate the posterior of the parametric density model of returns $\modelf{\theta}(\mcreturn|s)$ corresponding to the distributional agent from Step 1. Combining the standard Bayesian approach with the Bernstein-von Mises theorem, we will obtain a Gaussian posterior with a scaled inverse Fisher information as a covariance matrix. Finally we will observe that the posterior is efficient to approximate using log-likelihood gradients.

\subsubsection{Bayesian Posterior Derivation}
The derivation follows the typical Bayesian approach where
$ p(\theta | \Data) \propto p(\Data|\theta) p(\theta) $. Our data $\Data$ is comprised of $n$ state-return pairs $(\mcreturn_i, S_i)$ originating from a policy $\mu$. We denote the set of all $n$ observed returns as $\mcreturnseq$ and states as $\stateseq$. Akin to Bayesian regression we estimate the distribution of returns given input states:
\begin{align}
    p(\theta | \mcreturnseq, \stateseq)
    &\propto p(\mcreturnseq | \stateseq, \theta) p(\theta) \nonumber \\
    &\approx \prod_i
    \modelf{\theta}(\mcreturn_i| S_i) p(\theta) \label{eq:likelihoodfactorization}
\end{align}
In Bayesian regression the last step would be exact because the likelihood can be factorized $p(\mcreturnseq | \stateseq, \theta) = \prod_i \modelf{\theta}(\mcreturn_i | S_i) $. This is only possible if the $\mcreturn_i$ are conditionally independent given states, which is not the case for Monte-Carlo returns that are computed from overlapping reward sequences. Independence can be achieved when modelled $\bootstrapn$-step returns are used because all $\mcreturn'_i \sim f_\theta(\mcreturn | S_i)$ are conditionally independent given $\stateseq$. Return overlap is then reduced to $\bootstrapn$-steps, enabling two approaches: (a) sub-sampling the data to each $(\bootstrapn+1)^{\mathrm{th}}$ state will yield full independence and make Equation~\eqref{eq:likelihoodfactorization} exact (b) treating the factorization as an approximation that becomes more and more exact with smaller $\bootstrapn$. In the theoretical derivations we will assume independence e.g. achieved via (a). Incidentally such decorrelation via sub-sampling improves asymptotic agent performance in model based RL~\cite{Schrittwieser:2020}. In Algorithm~\ref{alg:epistemic_Qlearning} we chose (b) for simplicity and counter the potential overconfidence by rescaling the number of effective samples $n$ in our posterior approximation by a factor $\explorationscale$. 

\subsubsection{Approach to Posterior Approximation}
For large models the exact posterior is intractable and needs to be approximated. Following the Bernstein-von Mises theorem (which is explained below) we represent it as a Gaussian centered around $\thetamle$
\begin{equation}
    \label{eq:bvm_approximation}
    p(\theta|\Data) \approx \Normal{\thetamle, \frac{1}{n}\Fisher{\thetatrue}^{-1}}
\end{equation}
with covariance proportional to the inverse Fisher information matrix (which can be approximated as explained below):
$$
\Fisher{\thetatrue} 
    \defeq \Expectation[X\sim \modelf{\thetatrue}]{\gradtheta \log \modelf{\thetatrue}(X) \trans{\gradtheta \log \modelf{\thetatrue}(X)}} 
$$
A similar representation can be derived using the Laplace approximation (see appendix). The Gaussian structure permits efficient sampling.

\paragraph{Bernstein-von Mises Theorem}
The Bernstein-von Mises theorem states that the Bayesian posterior of a parametric density model $\modelf{\theta}(X)$ inferred from samples
$X_i \sim \modelf{\thetatrue}(X)$ from the true distribution
becomes a Gaussian distribution:
\begin{equation}
    p(\theta|X_1, ..., X_n) \to \normal{\thetamlen, \frac{1}{n}\Fisher{\thetatrue}^{-1}}
\end{equation}
Note that the Gaussian distribution is centered at the maximum likelihood solution $\thetamlen$. The covariance depends on the Fisher at the unknown true distribution parameters $\thetatrue$.
Observe that the covariance shrinks with the number $n$ of observed samples i.e. that the posterior gets narrower with $1/\sqrt{n}$ -- a property resembling the central limit theorem. Note that we slightly abuse notation since both sides in the limit depend on $n$.
More precisely stated the total variation norm between both distributions converges in probability to zero $\|p(\theta|X_1, .... X_n) - N(\thetamlen, \frac{1}{n}\Fisher{\thetatrue}^{-1}) \|_{TV} \to 0$. 
For a short summary please consider the appendix; for a detailed exposition please consider \citet{vandervaart1998:asymptotic,lecam1986:asympoticmethods,nickl2012:statistical} -- who date the theorem's origins back to \citet{Laplace:1810} work on the central limit theorem and early work by \citet{vonMises:1931} and \citet{Bernstein:TheoryProbabilities}.

The Bernstein-von Mises theorem relates Bayesian and frequentist statistics and is typically used to show that $||\thetamapn - \thetamlen|| \to 0$ and to argue that the prior distribution does not matter in the limit~\citep{vandervaart1998:asymptotic}. While not all theoretical assumptions can be satisfied in reinforcement learning with neural networks we observe favourable empirical results when employing it within the EVE recipe for value uncertainty estimation and exploration in Section~\ref{sec:experiments}.

\paragraph{Fisher Approximation}
We have $\thetamle$ from Step 1, but we are missing the Fisher information matrix $\Fisher{\thetatrue}$ to 
employ the Bernstein-von Mises theorem. 
Its expectation can be computed using the observed samples $X_i \sim \modelf{\thetatrue}$:
$
\Fisher{\thetatrue} 
    \approx \frac{1}{n} \sum_{i=1}^n \gradtheta \log \modelf{\thetatrue}(X_i) \trans{\gradtheta \log \modelf{\thetatrue}(X_i)} 
$
However it needs a further approximation because 
we can not compute gradients at the unknown $\thetatrue$.
\begin{align*}
    \EFisher{\thetamle}
    &\defeq \frac{1}{n} \sum_{i=1}^n \gradtheta \log \modelf{\thetamle}(X_i) \trans{\gradtheta \log \modelf{\thetamle}(X_i)}
\end{align*}
Now the gradients can be computed using automatic differentiation frameworks. Unfortunately estimating the full empirical Fisher information matrix $\EFisher{\thetamle}$ is infeasible with large-scale function approximation due to its quadratic memory requirements. However it can be efficiently approximated and inverted using a diagonal or Kronecker-factored representation \citep{Martens:2014}.

\subsection{Step 3: Epistemic Value Uncertainty}
Using Step 2 we can now efficiently sample parameters $\thetasample$ from our approximation to the posterior $p(\theta | \Data)$. We can use them for Thomson Sampling in parameter space, or to estimate the epistemic uncertainty of values. While $p(\theta | \Data)$ has a convenient Gaussian shape the nonlinearities in the model prevent us from deriving a similar analytic representation for $p(v_\theta(s)| \Data)$. Sampling epistemic values via $v'(s) \sim p(v_\theta(s)| \Data)$ is however easy and can for instance be used to estimate $\Variance{v_\theta(s)| \Data}$ numerically.
Sampling of $v'(s)$ can be achieved though sampling $\thetasample$ and evaluating $v'(s)\defeq v_\thetasample(s)$ which is the mean of the predicted return distribution $\modelf{\thetasample}(\mcreturn | s)$:
$v_{\thetasample}(s) = \int_\mcreturn \mcreturn \modelf{\thetasample}(\mcreturn | s) d\mcreturn$.
For Gaussian return models the predicted mean is the output of the networks forward pass and does not require integration.

\begin{algorithm}[tb]
\caption{\textbf{Standard Q-Learning} with $\epsilon$-greedy exploration.}
\label{alg:eps_Qlearning}
\textbf{Exploration Parameters:}
    Epsilon-greedy $\epsilon$.
    \\
\textbf{Regular Parameters:}
    Learning rate $\alpha$, neural network $q_\theta$, discount $\gamma$.
    \\
\textbf{Initialization:} Vector $\theta$ random. \\
\textbf{Acting:}
\begin{algorithmic}[1]
\State Play each action as $\argmax_a q_{\theta}(s, a)$ with probability $1-\epsilon$ or uniformly random otherwise.
\State Add resulting trajectory $\trajectory$ into experience replay $\mathcal{D}$.
\end{algorithmic}
\textbf{Q-Learning Update with $\theta$:}
\begin{algorithmic}[1]
\State Sample one transition $S_t, A_t, R_{t+1}, S_{t+1}$ from $\mathcal{D}$
\State $\return = R_{t+1} + \gamma \max_a q_{\theta}(S_{t+1}, a)$
\State $\theta \gets \theta - \alpha \gradtheta (\return - q_\theta(S_t, A_t))^2$
\end{algorithmic}
\end{algorithm}

\begin{algorithm}[tb]
\caption{\textbf{Epistemic Q-Learning using EVE} with diagonal Fisher approximation.}
\label{alg:epistemic_Qlearning}
\textbf{Exploration Parameters:}
     Exploration scale $ \explorationscale $, return variance ${\sigmareturn}^2$, 
    Fisher learning rate $\beta$, Fisher regularization $\epsilon$.
    \\
\textbf{Regular Parameters:}
    Learning rate $\alpha$, neural network $q_\theta$, discount $\gamma$.
    \\
\textbf{Initialization:} Vectors $\theta$ random, $\Fdiag$ zero. Scalar $n$ one. \\
\textbf{Acting:}
\begin{algorithmic}[1]
\State Sample $\thetasample$ from posterior.
\State Play one episode with $\argmax_a q_{\theta'}(s, a)$.
\State Add resulting trajectory $\trajectory$ into experience replay $\mathcal{D}$.
\State $n\gets n + |\trajectory|$
\end{algorithmic}
\textbf{Learning Step:}
\begin{algorithmic}[1]
\State Sample $\thetasample$ from posterior.
\State \textbf{Q-Learning Update with $\thetasample$:}
\State \ \ \ Sample one transition $S_t, A_t, R_{t+1}, S_{t+1}$ from $\mathcal{D}$.
\State \ \ \  $\return = R_{t+1} + \gamma \max_a q_{\theta'}(S_{t+1}, a)$ \label{alg:epistemic_Qlearning_ts_update}
\State \ \ \ $\theta \gets \theta - \alpha \gradtheta (\return - q_\theta(S_t, A_t))^2$
\State \textbf{Fisher Update:}
\State \ \ \ $\eta' \sim \normal{0, \sigmareturn}$
\State \ \ \ $\mcreturn' = \return + \gamma \eta'$
\State \ \ \  $\gradll = \gradtheta (\mcreturn' - q_\theta(S_t, A_t))^2$  
\State \ \ \ $\Fdiag \gets (1-\beta) \Fdiag  + \beta \gradll \odot \gradll$ \label{alg:epistemic_Qlearning_fisher_update}
\end{algorithmic}
\textbf{Posterior Sampling}:
\begin{algorithmic}[1]
\State Define the vectors  $\sigma$, $z$ such that for all $i$:
\State $\sigma_i = \frac{1}{\sqrt{(\Fdiag)_i + \epsilon}} $
\State Sample $z$ such that $z_i \sim \normal{0, 1}$.
\State \textbf{Return} $\theta + \frac{1}{\sqrt{n \explorationscale}} \sigma \odot z$
\end{algorithmic}
\end{algorithm}

\section{Epistemic Q-Learning}
\label{sec:epistemic_Qlearning}
The EVE recipe strives to provide deep RL agents with epistemic uncertainty estimates to improve their exploration. To provide a concrete example we use the recipe to convert a standard Q-Learning agent from Algorithm~\ref{alg:eps_Qlearning} into an illustrative epistemic Q-Learning agent  (Algorithm~\ref{alg:epistemic_Qlearning}).

In this section we strive for clarity over performance, hence where possible we prefer conceptually simpler approximations.
As a result we managed to keep the difference minimal: the Q-Learning update is modified slightly and the Fisher estimation is added, which is implemented with an exponential average of squared gradients from a noisy Q-Learning loss. These changes result in a computational overhead (one additional gradient pass) that is moderate compared to ensemble methods that store and update multiple model copies.
We discuss more advanced techniques for EVE such as distributional value functions, K-FAC approximations and variance reduction in the appendix. Nevertheless in Section~\ref{sec:experiments} we already observe competitive results with this illustrative agent on common benchmarks.

\paragraph{Standard Q-Learning Baseline}
The standard deep Q-Learning in Algorithm~\ref{alg:eps_Qlearning} uses a neural network $\qtheta$ to predict Q-values. The parameters $\theta$ are updated to minimize the squared prediction error towards targets $\return_{t}=R_{t+1} + \gamma \max_a q_\theta(S_{t+1}, a)$:
\begin{align}
    \label{eq:qlearninggrad}
    \loss^{Q-Prediction}(\theta) &\defeq (\return_{t} - q_\theta(S_t, A_t))^2
\end{align}
Note that the targets $\return_{t}$ are fixed and $\gradtheta\return_{t}=0$ (sometimes this is referred to as \emph{stop gradient}). Q-Learning then acts with $\epsilon$-greedy.

\paragraph{Introducing Thomson Sampling}
The first difference in Algorithm~\ref{alg:epistemic_Qlearning} is the introduction of Thomson Sampling at acting time. Furthermore we Thomson-sample at bootstrapping time by changing the target in line~\ref{alg:epistemic_Qlearning_ts_update} of the learning step to:
\begin{align}
    \return_{t}^{TS} & \defeq R_{t+1} + \gamma \max_a q_{\thetasample}(S_{t+1}, a) \\
    \thetasample &\sim p(\theta|\Data) \nonumber
\end{align}

\subsection{Applying Step 1: Model \& Log-Likelihood}
\label{sec:applying_step_1}
In this step we need to make the agent distributional and obtain the corresponding log-likelihood gradient. Rather than predicting the expected return $\qthetasa \approx \Expectation{G|s , a}$ 
it needs to predict the entire distribution of Monte-Carlo returns $\modelf{\theta}(\mcreturn|s, a) \approx p(\mcreturn | s, a)$.
Furthermore we require the ability to sample Monte-Carlo returns from the model: $\mcreturn' \sim \modelf{\theta}$.

\subsubsection{A Distributional Interpretation for Standard Q-Learning}
To make the example agent distributional and at the same time minimize algorithmic differences we simply reinterpret Q-Learning as a Gaussian density model which represents the Monte-Carlo return distribution as a Gaussian with fixed variance centered around the predicted Q-values $\qtheta$:
\begin{equation}
    \modelf{\theta}(\mcreturn | s, a) = \normal{\mcreturn | \mu=\qthetasa, \sigma=1} \label{eq:gaussian_model}
\end{equation}
enabling us to use the same powerful neural network architecture $\qtheta$ as the Q-Learning baseline. This algorithmically convenient choice of $\modelf{\theta}$ yields a powerful predictor of mean values but a crude return density model as it disregards state-dependent differences in the return variance. While this leaves room for future work, we will see in Figure~\ref{fig:n_vs_std} that it is able to capture state dependent epistemic value uncertainty and exhibits favourable exploration performance in our experiments in Section~\ref{sec:experiments}.

\subsubsection{Computing the Log-Likelihood Gradient}
The Gaussian model choice from Equation~\eqref{eq:gaussian_model}
implies that sampling bootstrap returns from $\modelf{\theta}(\cdot|S_{t+1}, a)$ simplifies to $\mcreturn' = q_{\theta}(S_{t+1}, a) + \eta'$ with $\eta'\sim \normal{0, 1}$. 
We can use that to derive the log-likelihood gradient for the model $\modelf{\theta}$ when it aims to predict the modelled return samples $\mcreturn'$ of following the Thomson Sampling policy:
\begin{align}
    \gradllt{\theta}
    \label{eq:grad_logl}
    &\defeq \gradtheta \log \modelf{\theta}(\mcreturn_{t}' |S_t, A_t) \\
    &=\gradtheta \frac{1}{2} (\mcreturn_{t}' -  q_\theta(S_t, A_t))^2 \nonumber
\end{align}
with a $\modelf{\theta}$-modelled Monte-Carlo return sample
\begin{align*}
    \mcreturn_{t}' &= \underbrace{R_{t+1} + \gamma \max_a q_{\thetasample}(S_{t+1}, a)}_{\return_{t}^{TS}} + \gamma \eta' \\
    \thetasample &\sim p(\theta|\Data) \text{\ , \ } \eta' \sim \normal{0, 1}
\end{align*}

\subsection{Applying Step 2: Posterior Approximation}
According to the recipe we approximate the posterior as
$p(\theta|\Data) \approx \normal{\thetamle, \frac{1}{n}\EFisher{\thetamle} ^{-1}}$
with 
\begin{align*}
\EFisher{\thetamle} 
    &\defeq \frac{1}{n} \sum_{t=1}^n \Expectation[\eta]{  \gradtheta \log \modelf{\thetamle}(Z_t') \trans{\gradtheta \log \modelf{\thetamle}(Z_t')} } \\
    &\defeq \frac{1}{n} \sum_{t=1}^n \Expectation[\eta]{ \gradll_t \ \ \trans{\gradll_t} }
\end{align*}
where $\gradll_t$ is short notation for the log-likelihood gradient from Equation~\eqref{eq:grad_logl} at the MLE estimate $\gradllt{\thetamle}$. 

Unfortunately estimating the full empirical Fisher information matrix is infeasible with large-scale function approximation due to its quadratic memory requirements. However it be efficiently approximated and inverted using diagonal or K-FAC representation~\citep{Martens:2014}. Striving again for the most simple example agent in this section we use the diagonal approximation:
$$  \hat{\mathcal{I}}^{\mathrm{Diag}}(\thetamle) \defeq \frac{1}{n} \sum_{i=1}^n \Expectation[\eta]{\gradll \odot \gradll} $$

\subsection{Applying Step 3: Epistemic Value Uncertainty}
Now estimation, inversion and sampling are efficiently possible using element-wise operations. Given a vector from a standard Normal $z \sim \normal{0, I}$ we can obtain a sample from:
$$ \thetasample \sim \normal{\thetamle, \frac{1}{n} \hat{\mathcal{I}}^{\mathrm{Diag}}(\thetamle)} \approx p(\theta| \Data)$$
by computing each component in the vector as
$ \thetasample_i \defeq (\thetamle)_i + \frac{z_i}{\sqrt{n \hat{\mathcal{I}}^{\mathrm{Diag}}(\thetamle)_i}} $.

\subsection{Optional Variance Reduction}
\label{sec:optional_variance_reduction}
The gradient 
$\gradllt{\theta} = \gradtheta \frac{1}{2}(\mcreturn_{t}' -  q_\theta(S_t, A_t))^2$
is stochastic because of the random variable $\eta'$ in 
$\mcreturn_{t}' = R_{t+1} + \gamma \max_a q_{\thetasample}(S_{t+1}, a) + \gamma \eta'$.
We can make our updates more sample-efficient by considering the corresponding expected updates. This is straightforward for the $\thetamle$ maximum likelihood parameter estimation where we recover the classic Q-Learning update:
\begin{align}
    \label{eq:expected_mle_grad}
    \gradmle_t(\theta) &\defeq \Expectation[\eta]{\gradllt{\theta}} \\
    &= (\Expectation[\eta]{\mcreturn_{t}'} -  q_\theta(S_t, A_t)) \gradtheta q_\theta(S_t, A_t) \nonumber\\
    &= (\return_t^{TS} -  q_\theta(S_t, A_t)) \gradtheta q_\theta(S_t, A_t) \nonumber\\
    &= \gradtheta \frac{1}{2}(\return_t^{TS} - q_\theta(S_t, A_t))^2 \nonumber
\end{align}
The variance of the Fisher update from line~\ref{alg:epistemic_Qlearning_fisher_update} in Algorithm~\ref{alg:epistemic_Qlearning} could also be reduced, but this is non-trivial (see appendix).

\section{Related work}
How to measure epistemic uncertainty in deep reinforcement learning is actively researched. Popular methods involve ensembles consisting of multiple independent neural networks
or sharing parts of them \citep{Osband:2016} resulting in increased memory and computational requirements. Heuristics, like used by~\citet{burda2018exploration}, use the prediction error of an unknown random function target as an exploration bonus. \citet{ostrovski2017count} employed a generative model of states to compute pseudo-counts for exploration. \citet{Dearden:1998} approximated the posterior of state-action values for exploration in tabular MDPs. In non-tabular MDPs~\citet{Deisenroth:2011} employed techniques from kernel methods resulting in increased data-efficiency but also being limited by the large computational cost of kernel regression. Similarly ~\citet{chua2018:handful,curi2020:huclr} approximate the environment dynamics with an ensemble of neural networks each parameterizing the next states probability with a Gaussian.
In the field of Bayesian deep learning \citep[cf.][for early discussions of the principles]{MacKay:92b,Hinton1993:descriptionlenght,Neal1995:Bayesian} a series of innovations were proposed  \citep{Graves2011:variational,blundell2015:BBB,Gal2016:dropout,Korattikara2015:darkknowledge,Lakshminarayanan2017:ensembles,Osband2018randomized,Zhang2018:noisynaturalgrad,ritter2018:laplace,daxberger2021:laplace} \citep[see][for an overview]{Murphy2023:pbml}. However there is no consensus of how to adapt from supervised learning to sequential decision-making processes, where value-bootstrapping and off-policy learning need to be considered. \citet{tang2018:exploration} use distributional reinforcement learning~\cite{Bellemare2017distributional} for exploration and provide a variational interpretation for \citet{plappert2018:parameter} and \citet{fortunato2018:noisy} that perturb the parameters of reinforcement learning agents with heuristically scaled Gaussian noise. In our work we derive an explicit approximation to the posterior that is motivated by the Bernstein von-Mises theorem from statistical theory. The posterior considers all agent parameters distinguishing it from last-layer methods such as \citet{Azizzadenesheli2018:bayesian}. In particular it is compatible with neural network function approximation, efficient to estimate (using automatic differentiation methods) and efficient to evaluate. Compared to ensembles that maintain and update multiple models it requires fewer parameters and typically less compute. Furthermore it supports unlimited sampling of values, where ensembles only provide a constant number of values (one sample per parameter copy).

\section{Experiments}
\label{sec:experiments}
\begin{figure}[t]
\centering
\includegraphics[width=0.45\textwidth]{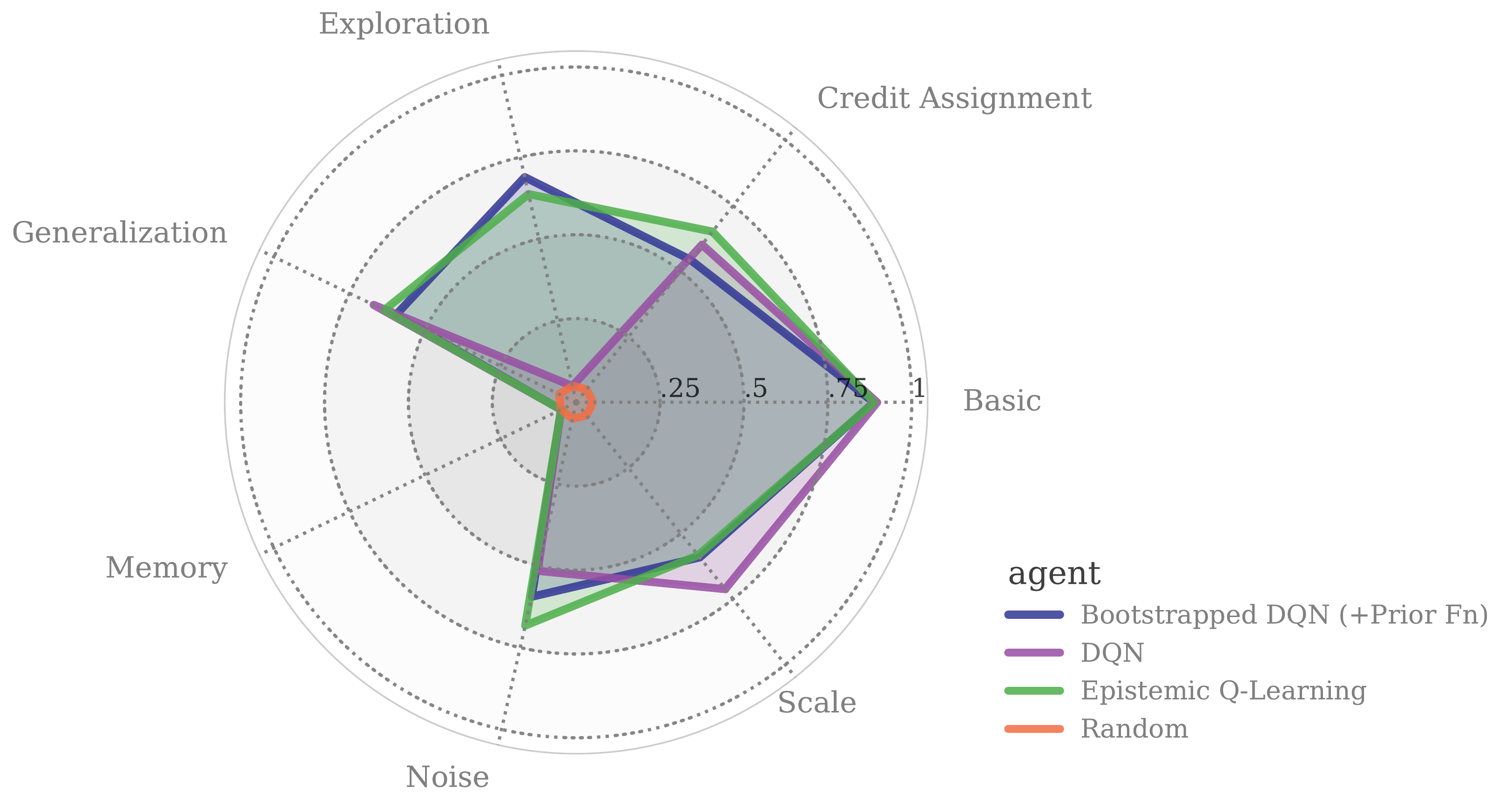}
\caption{Performance breakdown of selected agents along the Bsuite capabilities. Observe that Epistemic Q-Learning and Bootstrapped DQN achieve high exploration scores while regular DQN barely exceeds a random agent. The score breakdown is aggregated from 468 Bsuite task variations averaged over 30 holdout seeds.}
\label{fig:spider}
\end{figure}

We have proposed a general recipe for epistemic value estimation (EVE), derived a simple example agent from it in Section~\ref{sec:epistemic_Qlearning} and empirically evaluate it here. In Figure~\ref{fig:spider} we observe competitive performance on the Bsuite benchmarks~\citep{osband2020bsuite}, where our agent matches the state-of-the-art results from~\citet{Osband2018randomized} that employs an ensemble of 20 independent neural network copies each with their own copy of a random prior function, target network and optimizer state. In comparison epistemic Q-Learning with EVE requires only a single copy of network, target network, diagonal fisher and optimizer state -- resulting in $20\times$ fewer parameters.
Furthermore we present ablations  (Figure~\ref{fig:ablations}) and parameter studies (Figure~\ref{fig:studies}) showing that our agent is robust to the choice of hyper-parameters.

\subsection{Benchmark Environments}

We use the behaviour suite benchmark (Bsuite) with special focus on the Deep Sea environment to empirically analyze our epistemic Q-Learning example agent from Section~\ref{sec:epistemic_Qlearning}.

\paragraph{Behaviour Suite}
Bsuite was introduced by~\citet{osband2020bsuite} to facilitate the comparison of agents not just in terms of total score but across meaningful capabilities (such as exploration, credit assignment, memory, and generalization with function approximators). The standardized evaluation protocol facilitates direct comparisons between research papers. 
Bsuite consists of 13 environments (including versions of Deep Sea, Mountaincar, Cartpole, contextual bandits with MNIST images and T-maze environments) which are then varied in difficulty (such as problem size, reward sparsity, or stochasticity of transitions or reward) resulting in 468 task variations. Each agent is evaluated on all 468 tasks and the performance is grouped into 7 categories (called capabilities). We are most interested in the \emph{exploration capability score} which considers the sparse reward tasks (Deep Sea, Stochastic Deep Sea and Cartpole Swingup) with various difficulty levels resulting in 62
task variations. The aggregate exploration score is the fraction of those 62 tasks where the sparse reward is consistently\footnote{
In line with prior publications we use a harder evaluation metric than originally described in~\citet{osband2020bsuite} -- see Appendix~\ref{sec:misc_deep_sea_threshold}.
} obtained by the agent.

\begin{figure}[t]
\centering
\includegraphics[width=0.20\textwidth]{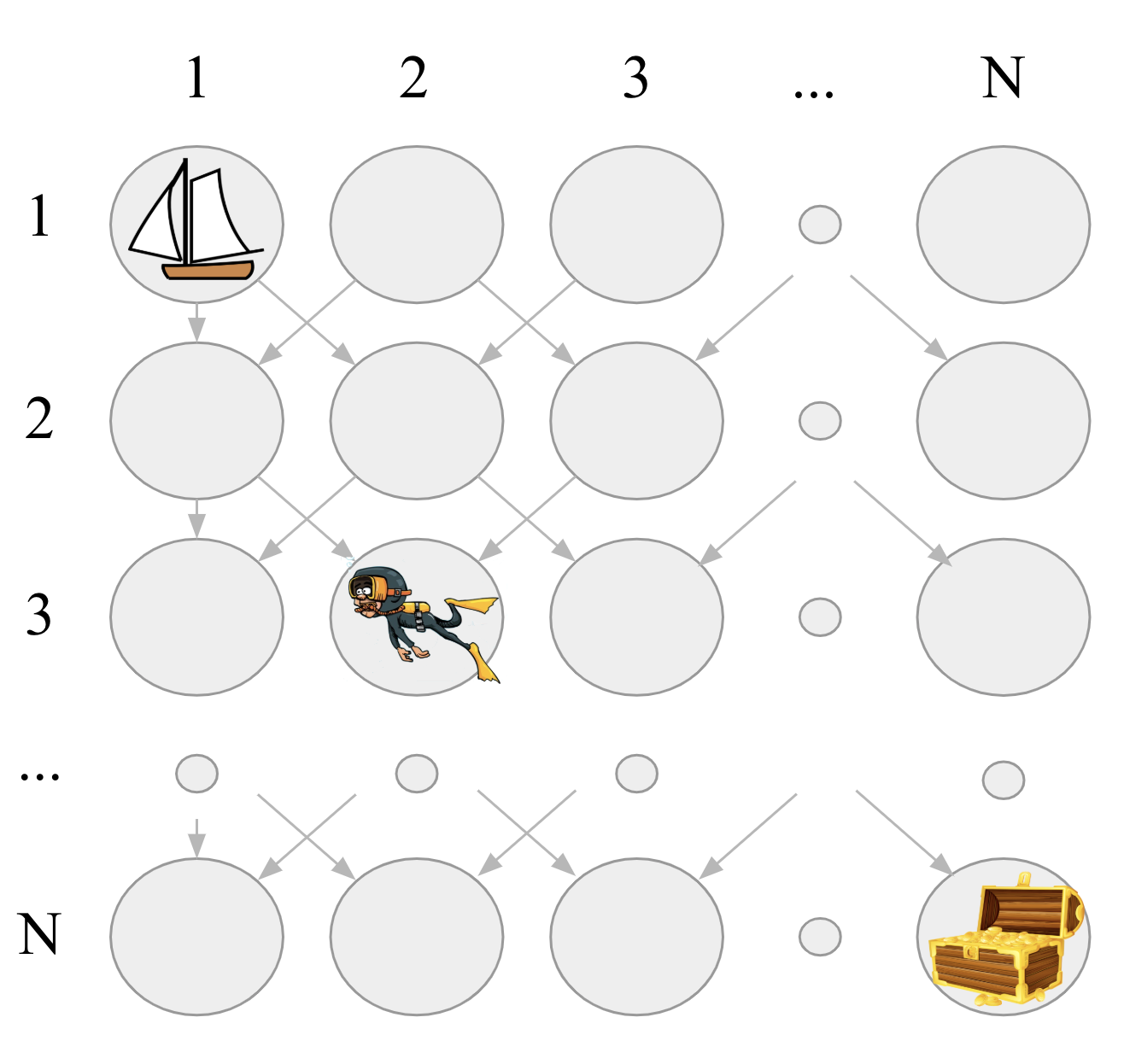}
\caption{The Deep Sea MDP is given by an $L$ by $L$ sized grid, where a diver starts in the top left corner and may move down right or down left at each step. A treasure is on the far right side, but the agent can only reach it by moving right at all steps. Hence only a single  among $2^{L-1}$ possible action sequences is successful. Trivial solutions are prevented by randomizing the action mapping. To discourage the correct sequence there is a penalty for each movement to the right.}
\label{fig:deep_sea_image}
\end{figure}

\begin{algorithm}[t]
\caption{\textbf{Epistemic Q-Learning using EVE} with diagonal Fisher approximation, burn-in, target networks, and ADAM optimization (omitting mini-batching details).}
\label{alg:full_epistemic_Qlearning}
\textbf{Exploration Parameters:}
     Exploration scale $ \explorationscale $, return variance ${\sigmareturn}^2$, 
    Fisher learning rate $\beta$, Fisher regularization $\epsilon$, burn-in steps $K_{\mathrm{burnin}}$.
    \\
\textbf{Regular Parameters:}
    Learning rate $\alpha$, target network period $K_{\mathrm{target}}$, neural network $q_\theta$, discount $\gamma$.
    \\
\textbf{Initialization:}
Vectors $\theta, \thetatarget$ random, $\Fdiag$ zero. Scalars $m, n$ one. \\
\textbf{Acting:}
\begin{algorithmic}[1]
\State Act uniformly for the first $K_{\mathrm{burnin}}$ episodes; otherwise:
\State Sample $\thetasample$ from posterior.
\State Play one episode with $\argmax_a q_{\theta'}(s, a)$.
\State Add resulting trajectory $\trajectory$ into experience replay $\mathcal{D}$.
\State $n\gets n + |\trajectory|$
\end{algorithmic}
\textbf{Learning Step:}
\begin{algorithmic}[1]
\State Sample $\thetasample$ from posterior.
\State \textbf{Q-Learning Update with $\thetasample$:}
\State \ \ \ Sample one transition $S_t, A_t, R_{t+1}, S_{t+1}$ from $\mathcal{D}$.
\State \ \ \  $\return = R_{t+1} + \gamma \max_a q_{\theta'}(S_{t+1}, a)$ \label{alg:full_epistemic_Qlearning_distributional_update}
\State \ \ \  $\gradmle = \gradtheta (\return - q_\theta(S_t, A_t))^2$ 
\State \ \ \ $\theta \gets \theta - \alpha \, \mathrm{ADAM}(\gradmle)$
\State \textbf{Fisher Update:} 
\State \ \ \ $\eta' \sim \normal{0, \sigmareturn}$
\State \ \ \ $\mcreturn' = \return + \gamma \eta'$
\State \ \ \  $\gradll = \gradtheta (\mcreturn' - q_\theta(S_t, A_t))^2$  
\State \ \ \ $\Fdiag \gets (1-\beta) \Fdiag  + \gradll \odot \gradll$ \label{alg:full_epistemic_Qlearning_fisher_update}
\State \ \ \ $m \gets (1-\beta) m + 1$
\State Every $K_{\mathrm{target}}$ steps update the target network $\bar{\theta} \gets \theta$.
\end{algorithmic}
\textbf{Posterior Sampling}:
\begin{algorithmic}[1]
\State Define the vectors $\Funbiased$, $\sigma$, $z$ such that for all $i$:
\State $(\Funbiased)_i = (\Fdiag)_i / m$ 
\State $\sigma_i = \frac{1}{\sqrt{(\Funbiased)_i + \epsilon}} $
\State Sample $z$ such that $z_i \sim \normal{0, 1}$.
\State \textbf{Return} $\thetatarget + \frac{1}{\sqrt{n \explorationscale}} \sigma \odot z$
\end{algorithmic}
\end{algorithm}

\begin{figure}[t]
\centering
\includegraphics[width=0.35\textwidth]{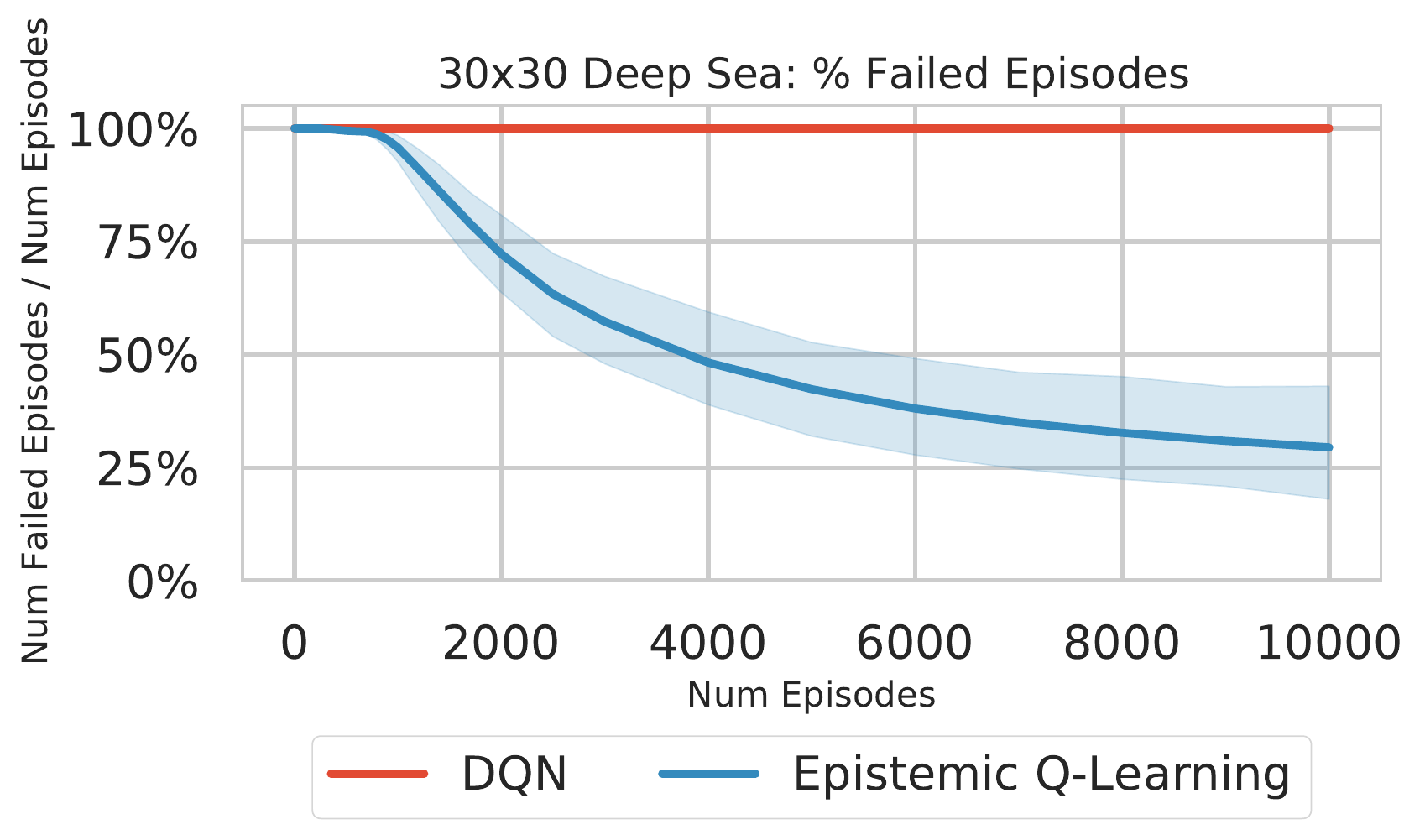}
\caption{Epistemic Q-Learning solves the Deep Sea ($30\times30$) exploration benchmark: the failures among the total number of episodes diminish. DQN fails to solve it.}
\label{fig:deep_sea_convergence}
\end{figure}

\paragraph{Deep Sea}
\citet{Osband2018randomized} propose the Deep Sea environment (see Figure~\ref{fig:deep_sea_image}) to measure exploration performance. It is a `needle in the haystack'-style problem where $\epsilon$-greedy requires an exponential number of environment steps because it over-explores. It was proposed to motivate the the need for `deep' or `directed' exploration with polynomial exploration time. 
Figure~\ref{fig:deep_sea_convergence} shows that DQN with its $\epsilon$-greedy exploration is unable to find the treasure, while Epistemic Q-Learning obtains the treasure in more than $70\%$ of the given 10000 episodes and $30$ seeds.

\subsection{Methodology}
\label{sec:method}
We build our agent on the reference agent implementations from Bsuite and strive to minimize all unrelated differences (e.g. neural network architecture, optimizer, target networks) to permit a clear comparison -- see appendix for algorithmic details. Most notably we replace ReLU with Leaky-ReLU activations as they yield non-zero gradients almost everywhere while maintaining the ReLU benefits and confirm that the choice of activation does not change the DQN baseline performance. Exploration hyper-parameters were tuned among possible powers of $10$ in the range $[10^{-15}, 10^{10}]$.
We then sampled new holdout seeds for all figures ($10$ holdout seeds for the parameter studies and $30$ for epistemic Q-Learning, all baselines and ablations). Hence a complete Bsuite evaluation in Figure~\ref{fig:spider} evaluates all $468$ tasks $30$ times with the exploration score aggregating $62 \times 30$ separate RL runs. Figures~\ref{fig:studies} and~\ref{fig:ablations} summarize $468 \times 10 \times 7 \times 4$ and respectively $62 \times 30 \times 5$ separate RL runs.

\subsection{Experimental Study on Bsuite}
In Figure~\ref{fig:spider} we compare to the DQN and Bootstrapped DQN reference implementations from Bsuite. The latter uses an ensemble of 20 agents combined with random prior functions~\citep{Osband2018randomized}. 
Our agent matches Bootstrapped DQN in important parameters such as network capacity, number of training steps and optimizer settings. We strive to minimize changes for better comparability -- see Algorithm~\ref{alg:full_epistemic_Qlearning} and appendix for implementation details. However note that Bootstrapped DQN maintains 20 separate agent copies in an ensemble increasing the number of parameters and computational requirements significantly.

\begin{figure}[t]
\centering
\includegraphics[width=0.45\textwidth]{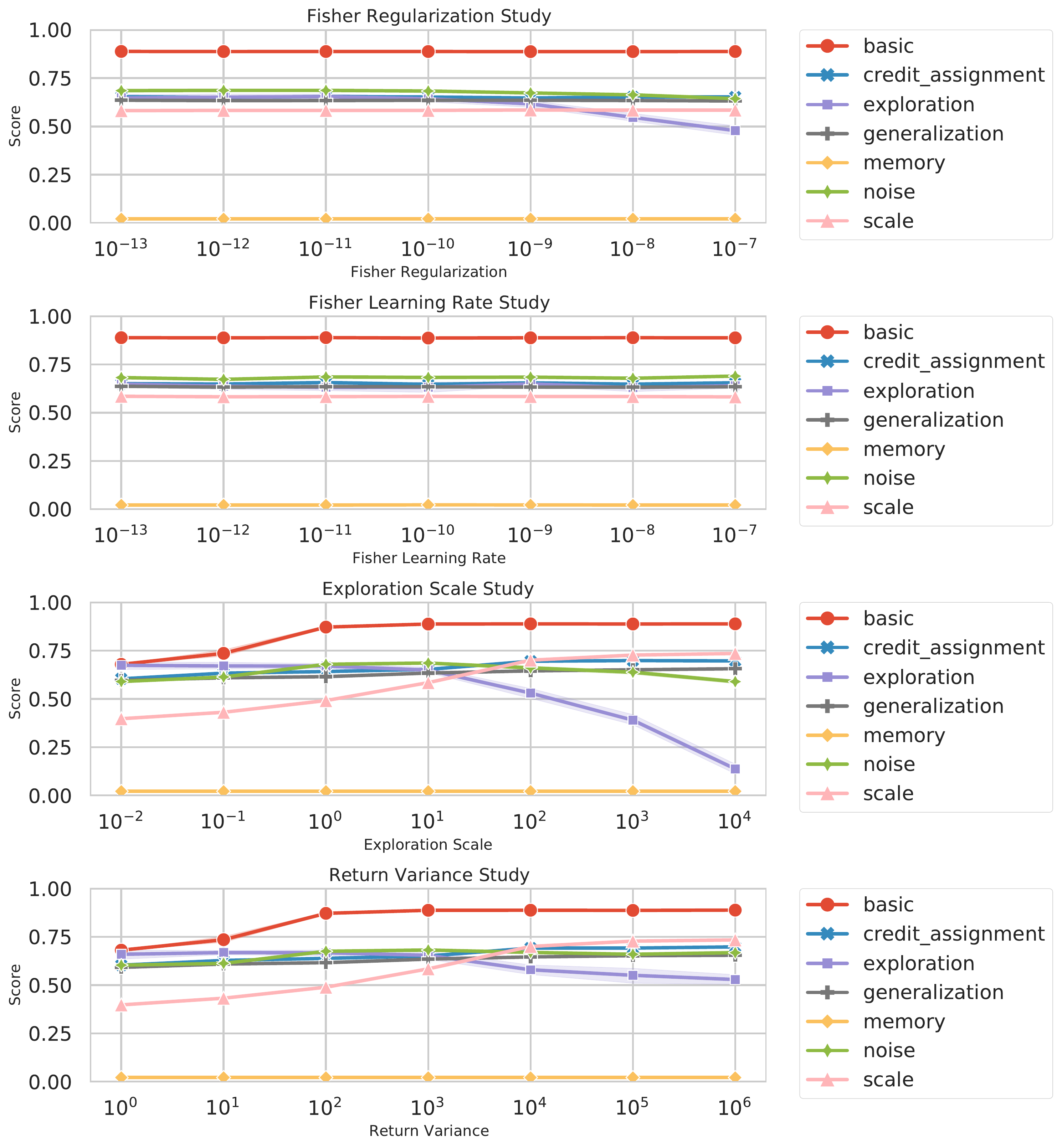}
\caption{Parameter sensitivity study of Epistemic Q-Learning broken down by Bsuite capability using 10 holdout seeds. Centered values on the x-axis are the default parameters. Observe that parameters are robust across multiple orders of magnitude. }
\label{fig:studies}
\end{figure}

Empirically we observe on-par performance of Epistemic Q-Learning and Bootstrapped DQN: Both score nearly equal across all dimensions. In particular they score high in the
exploration dimension, where regular DQN is barely better than the random agent. On the other hand they are both a bit worse than DQN in the scale dimension. Epistemic Q-Learning can however trade off the performance on scale vs. exploration by hyper parameter selection (see Figure~\ref{fig:studies}).

\begin{figure}[t]
\centering
\includegraphics[width=0.35\textwidth]{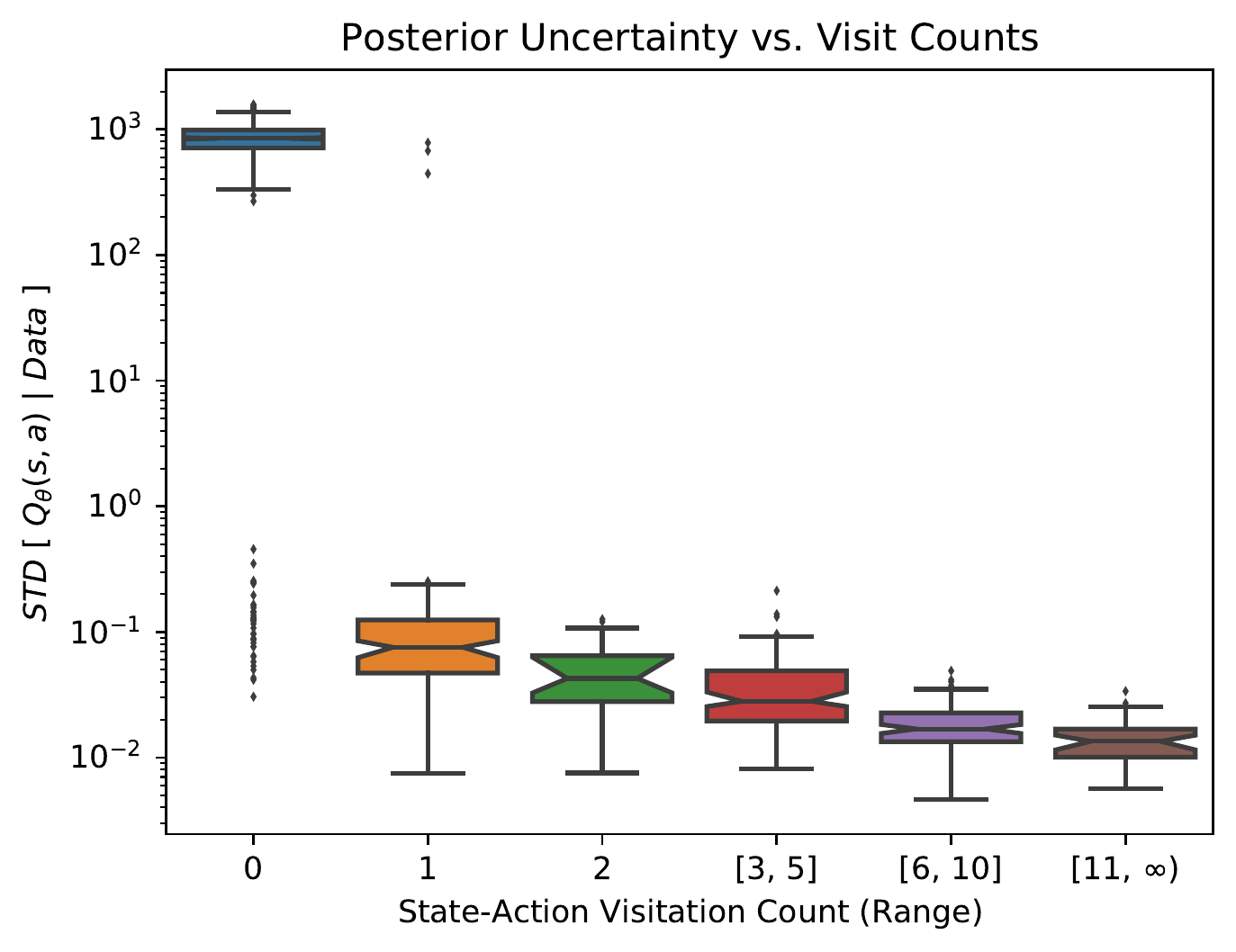}
\caption{As desired, the predicted epistemic uncertainty $\sqrt{\Variance{q_\theta(s, a)|\Data}}$ is lower at frequently visited and hence better explored states. Evaluated on a $30\times30$ Deep Sea.}
\label{fig:n_vs_std}
\end{figure}

\subsection{Probing for Epistemic Uncertainty}
In Figure~\ref{fig:n_vs_std} we ask the fundamental question whether our predicted epistemic uncertainty makes sense empirically: i.e. if uncertainty is high at unknown states and low at frequently visited states. 
We observe affirmative evidence from the following sanity check: we consider a $30\times30$ Deep Sea and collect 100 episodes with a uniformly random policy, while estimating epistemic Q-values using the Epistemic Q-Learning algorithm. In Figure~\ref{fig:n_vs_std} we compare the actual number of visits (x-axis) with the posterior standard deviation of the Q-values at each state (y-axis). To simplify the plot we group states into visit count ranges. One can observe a clear correlation where uncertainty roughly decreases the more often a state is visited. In particular unknown states with zero visitations exhibit a significantly larger uncertainty than visited states.

\subsection{Parameter Stability}
An ideal algorithm should be robust to hyper-parameters and consequently require minimal tuning. In Figure~\ref{fig:studies} we observe that Epistemic Q-Learning is robust with respect to all parameters: to both the Fisher regularization and Fisher learning rate across 5 orders of magnitude (from $10^{-13}$ to $10^{-8}$). The exploration scale and return variance parameters trade-off exploration and exploitation i.e. too large values seem to hinder exploration while improving the scale capability. Good trade-offs are achieved in the range of $1$ to $10^2$ for the exploration scale and $10^2$ to $10^6$ for the return variance.

\subsection{Ablations}
\label{sec:ablations}
\begin{figure}[h]
\centering
\includegraphics[width=0.45\textwidth]{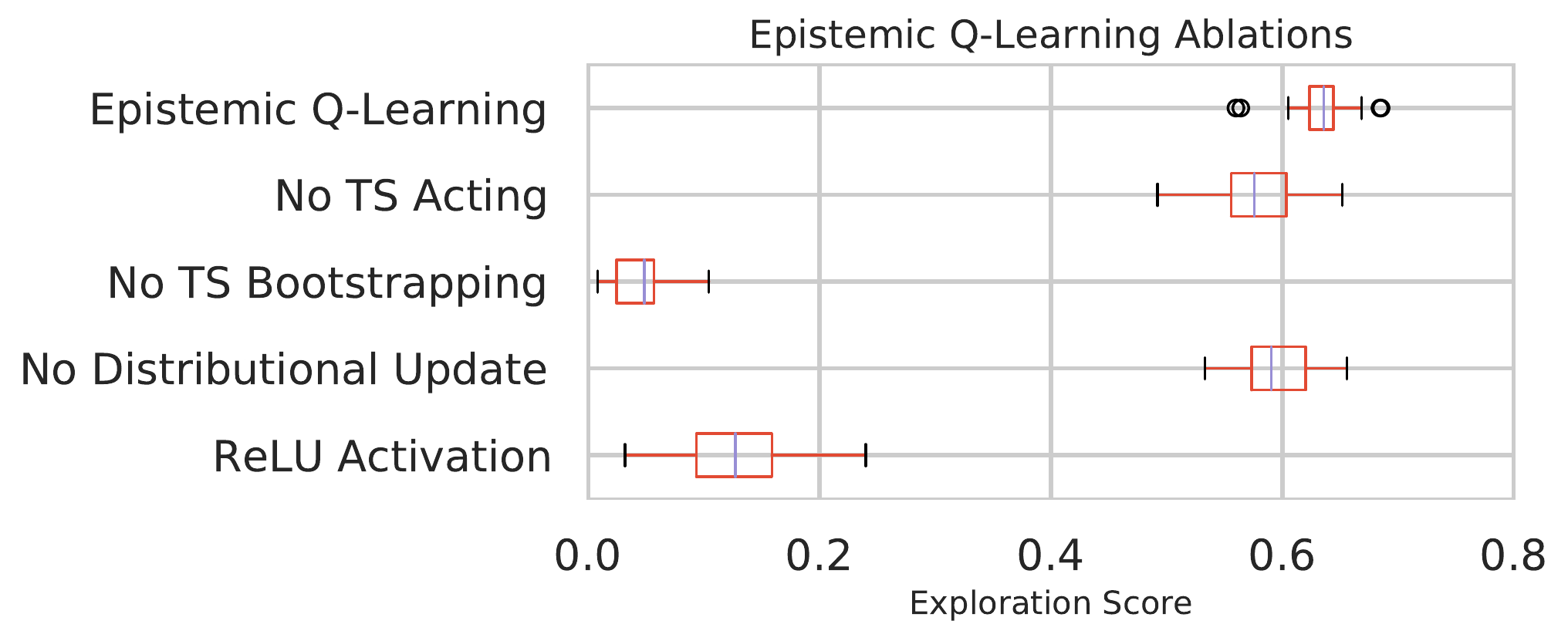}
\caption{Various ablations of Epistemic Q-Learning applied to Bsuite with 30 seeds (experimental details in Section~\ref{sec:ablations}).}
\label{fig:ablations}
\end{figure}

In Figure~\ref{fig:ablations} we ablate key components of the Epistemic Q-Learning agent (Algorithm~\ref{alg:full_epistemic_Qlearning}) and evaluate the corresponding Bsuite exploration score to answer the following questions:
\begin{itemize}
    \item \textbf{What happens if we replace the Thomson Sampling acting?}
    We replace it by $\epsilon$-greedy as in DQN. Note that we keep the Thomson-Sampling inspired bootstrapping as is. We observe a drop in performance.
    \item \textbf{What happens if we replace the Thomson-Sampling inspired value-bootstrapping?}
    We replace line~\ref{alg:full_epistemic_Qlearning_distributional_update} in the learning step of Algorithm~\ref{alg:full_epistemic_Qlearning} by a distributional update with the Q-function at $\thetamle$ by setting: $\mcreturn_{t} = R_{t+1} + \gamma \max_a q_{\thetamle}(S_{t+1}, a) + \gamma \eta'$. We observe a large drop in performance indicating that bootstrapping from a posterior sampled $q_{\thetasample}$ is important for exploration.
    \item \textbf{What happens if we remove the distributional update?} 
    When estimating the Fisher in Algorithm~\ref{alg:full_epistemic_Qlearning} line~\ref{alg:full_epistemic_Qlearning_fisher_update}
    with a regular non-distributional Q-Learning update $\gradmle_t$ we observe a drop in performance and increased variance.
    \item \textbf{What happens if we use ReLU activations?} We use Leaky-ReLU activations~\citep{Maas13rectifiernonlinearities,Xu2015activations} as they have a non-zero gradient almost everywhere. We observe a dramatic drop in performance if we use ReLU activations instead and hypothesize that this is due to the zero gradient it exhibits on all negative input values interacting with the posterior estimation. Note that the DQN-baseline performance does not improve with Leaky-ReLUs. Hence Leaky-ReLUs are a prerequisite for efficient exploration using EVE, but not its cause.
\end{itemize}

\section{Conclusion}
Motivated by statistical theory we introduced a principled recipe for posterior approximation (EVE) that is compatible with sequential decision making and with neural network function approximation. In practice it requires moderate computational and memory overhead and no architecture changes besides introducing Leaky-ReLU activations. We applied the recipe to compute epistemic value uncertainty in a Q-Learning agent and used it for Thomson-Sampling inspired exploration. The proposed epistemic Q-Learning agent exhibits competitive performance on a series of general reinforcement learning and dedicated exploration benchmarks. It is robust to hyper-parameters and matches the performance of Bootstrapped DQN on Bsuite with $20\times$ fewer agent parameters.

To facilitate understanding and analysis, the presented Epistemic Q-Learning agent was designed with a focus on simplicity. It can be extended in many ways: with more expressive distributional representations~\citep{Bellemare2017distributional}, more-sophisticated Fisher approximations~\citep{Martens:2014}, or better use of epistemic uncertainty in the policy construction~\citep{nikolov2018informationdirected}. While we employed the approximate posterior for exploration, future work may evaluate its benefits for other applications such as safety in reinforcement learning or robust offline learning.

\bibliography{aaai23}

\appendix
\section*{Appendix}

\section{Implementation Details}
\label{sec:implementation_details}
We build our agent on the reference agent implementations from Bsuite. In particular we strive to minimize all unrelated differences to Bootstrapped DQN to obtain a clear comparison. Notable differences are replacing ReLU with Leaky-ReLU activations and using 10 small batches of size 12 rather than one large batch of size 128 and increasing the replay capacity such that the agent can remember all previous transitions. In particular we maintain the neural network architecture. While Bootstrapped DQN ensembles 20 copies Epistemic Q-Learning uses a single copy.

Compared to the high-level example 
Algorithm~\ref{alg:epistemic_Qlearning}
in the paper it inherits a target network~\citep{Mnih:2015} and ADAM optimizer~\citep{Kingma:2015} with learning rate $10^{-3}$ from the Bsuite reference implementations. Furthermore it performs maximum likelihood optimization with the expected gradient 
(Equation~\ref{eq:expected_mle_grad}) to reduce variance. See Algorithm~\ref{alg:full_epistemic_Qlearning} for a full description of the agent modulo mini-batching. To simplify the mini-batching implementation we use a single posterior sample $\theta'$ to compute all returns in the batch $\return_t = R_{t+1} + \gamma \max_a q_{\theta'}(S_{t+1}, a)$. Furthermore we compute the gradient $\gradll$ by summing the prediction loss over all transitions batch $\gradtheta \sum_{t \in B} (\mcreturn_t' - q_\theta(S_t, A_t))^2$. A gold standard implementation would implement batching by sampling a different posterior sample $\theta'$ and computing the gradient $\gradll$ for each transition in the batch. We mitigate the resulting approximation error by using a smaller batch size compared to the baseline (12 rather than 128). Finally it acts uniformly random for the first $K_{\mathrm{burnin}}=100$ episodes to ensure there the Fisher information matrix has been estimated sufficiently before we sample.

\section{Variance Reduction of Fisher Estimation}
In Step 2 of the recipe we approximate the posterior as $p(\theta|\Data) \approx \normal{\thetamle, \frac{1}{n}\EFisher{\thetamle} ^{-1}}$. In
Section~\ref{sec:recipe}
we have discussed how to obtain the $\thetamle$ and the Fisher information matrix $\EFisher{\thetamle}$ by sampling hypothetical returns from the model $\mcreturn'\sim\modelf{\theta}$. Sampling introduces additional variance, which can however be removed for the Gaussian return model: We discussed variance reduction for the MLE estimation in Section~\ref{sec:optional_variance_reduction}
and will now focus on the Fisher information matrix:
\begin{align*}
\EFisher{\thetamle} 
    &\defeq \frac{1}{n} \sum_{t=1}^n \Expectation[Z_t']{  \gradtheta \log \modelf{\thetamle}(Z_t') \trans{\gradtheta \log \modelf{\thetamle}(Z_t')} } \\
    &\defeq \frac{1}{n} \sum_{t=1}^n \Expectation[Z_t']{ \gradll_t \ \ \trans{\gradll_t} }
\end{align*}
This estimator would become quite expensive if we implemented it naively by summing over $n$ transitions in the dataset and then evaluating each $\expectation[Z_t']{ \gradll_t \ \ \trans{\gradll_t} }$ numerically by sampling $Z_t'$ say $m\gg 1$ times. Below we present a method to compute the expectation analytically at similar computational cost as $m=1$. 

As the approach below is mathematically a bit more involved one might be tempted to instead use $\frac{1}{n} \sum_{t=1}^n\expectation[Z_t']{\gradll_t} \trans{\expectation[Z_t']{\gradll_t}}=\frac{1}{n} \sum_{t=1}^n\gradmle_t \trans{\gradmle_t}$. This is tempting as we already have $\gradmle_t$ from Section~\ref{sec:optional_variance_reduction}. Unfortunately this is however not theoretically justified because 
$\expectation[Z_t']{\gradll_t \trans{\gradll_t}} \neq \expectation[Z_t']{\gradll_t} \trans{\expectation[Z_t']{\gradll_t}}$ and empirically results in inferior performance (see Section~\ref{sec:ablations}, Figure~\ref{fig:ablations}: `No Distributional Update').

\paragraph{Fisher Variance Reduction for the Gaussian Model}
When employing a Gaussian model $\modelf{\theta}(\mcreturn) = \normal{\mcreturn | v_\theta, \sigma=1}$ with $\bootstrapn$-step returns $G^\bootstrapn$ we sample hypothetical returns as $\mcreturn' = G^\bootstrapn + \gamma^\bootstrapn \eta$ with $\eta \sim \normal{0, 1}$. The log-likelihood gradient then simplifies as follows:
\begin{align*}
    \gradllt{\theta}
    &\defeq \gradtheta \log \modelf{\theta}(\mcreturn_{t}' |S_t, A_t) \\
    &= \gradtheta \frac{1}{2} (\mcreturn_{t}' -  q_\theta(S_t, A_t))^2 \\
    &=  \frac{1}{2} (\mcreturn_{t}' -  q_\theta(S_t, A_t)) \gradtheta q_\theta(S_t, A_t) \\
    &=  \frac{1}{2} (\underbrace{G^\bootstrapn_t -  q_\theta(S_t, A_t)}_{\defeq \delta_t^{TD}} + \gamma^\bootstrapn \eta ) \gradtheta q_\theta(S_t, A_t)
\end{align*}
Hence we can pull the gradient computation out of the expectation: denoting $\delta_t^{TD} \defeq G^\bootstrapn_t -  q_\theta(S_t, A_t)$ and $\gradtheta \qtheta \defeq \gradtheta q_\theta(S_t, A_t) $
\begin{align*}
\Expectation[Z_t']{\gradllt{\theta} \trans{\gradllt{\theta}}} 
    &=   \Expectation[\eta]{ \frac{1}{4} (\delta_t^{TD} + \gamma^\bootstrapn \eta)^2 \gradtheta \qtheta \trans{\gradtheta \qtheta}} \\
    &=   \Expectation[\eta]{ \frac{1}{4} (\delta_t^{TD} + \gamma^\bootstrapn \eta)^2} \gradtheta \qtheta \trans{\gradtheta \qtheta} \\
    &=    \frac{1}{4}\left((\delta_t^{TD})^2 + \gamma^{2\bootstrapn}\right) \gradtheta \qtheta \trans{\gradtheta \qtheta} 
\end{align*}
using $\Expectation{X^2} = \Expectation{X}^2 + \Variance{X}$ in the last step we can compute the expectation analytically. Overall we can now compute the expected Fisher update for transition $t$ with a single gradient computation of $\gradtheta \qtheta$ and with $n$ computations for the entire dataset. Furthermore we can share computation with the MLE optimization since both employ the same gradient $\gradtheta \qtheta$.

\section{Posterior Approximation}
Exact posterior estimation is intractable in high dimensional parameter spaces hence approximations have been studied extensively \citep[see][for an overview]{Murphy2023:pbml}. Historically the Bernstein-von Mises theorem is an extension and improvement of previous Laplace approximation based attempts to estimate the posterior of a parametric model. 

\paragraph{Comparison of Bernstein-von Mises and Laplace Approximations}
Approximating the posterior with a Gaussian
$$ p(\theta|\Data) \approx \normal{\theta^*, \Sigma} $$
can be justified in two similar ways.
Inspired by the frequentist Bernstein-von Mises theorem:
$$ p(\theta|\Data) \to \normal{\thetamlen, \frac{1}{n}\Fisher{\thetatrue}^{-1}} $$
or using the Laplace approximation:
$$ p(\theta|\Data) \approx \normal{\thetamapn, \Hessian{\thetamapn}^{-1}} $$
Prior to~\citet{vonMises:1931} it was common to heuristically approximate posteriors using the Laplace formula without sound theoretical understanding of how accurate the approximation is. Hence \citet{vonMises:1931} emphasizes that his work provides the missing clarity by proving that the posterior distribution convergences to a Gaussian -- a result that he named \emph{the second law of large numbers}. While originally limited to specific statistical models the result has been extended by ~\citet{lecam1953:bernsteinVonMises} and referred to as \emph{Bernstein-von Mises theorem}.

While the means in both formulas appear different note that they are equal in the presence of regularization $\thetamlen=\thetamapn$. Now let us compare the covariance matrices: $\Fisher{\thetatrue}$ is the Fisher information at the unknown true parameters and needs to be approximated. $\Hessian{\thetamapn}$ is the Hessian of the negative log-posterior and both hard to estimate and not guaranteed to be positive semi-definite, hence impractical as a covariance matrix. Usually it is heuristically replaced by something more suitable. Two common options are the Fisher information matrix and the Generalized Gauss Newton~\citep{Murphy2023:pbml,Botev2017:GaussNewton,ritter2018:laplace}. \citet{Martens:2014} remarks that the Fisher and GGN matrix are equal for exponential families such as the Gaussian model used in our example agent from see Section~\ref{sec:applying_step_1}.
In Section~\ref{sec:experiments} we observed favourable results with the empirical Fisher matrix. Optimizing the matrix choice is out of scope for this paper. We refer to \citet{Martens:2014} and \citet{Murphy2023:pbml} for a discussion of the matrix choice and conclude that the posterior approximation in Step 2 can be justified both using the Bernstein-von Mises theorem and the Laplace approximation.

\paragraph{Bernstein-von Mises Theorem}
The Bernstein-von Mises theorem states that the posterior of a parametric model converges to a Gaussian distribution and that ultimately the prior does not matter. While it can be used to relate frequentist and Bayesian inference for large sample sizes, we use it simply for its parametric form.

At the core of frequentist statistics is the existence of an unknown true parameter $\thetatrue$ parameterizing the true distribution from which the samples are observed:
$$X_i \sim \modelf{\thetatrue}$$
The true distribution parameter $\thetatrue$ is then estimated from a set of $n$ samples -- e.g. using the maximum likelihood estimator:
\begin{align}
    \thetamlen \defeq \argmax_\theta \prod_{i=1}^n \modelf{\theta}(X_i)
\end{align}
The Bernstein-von Mises theorem states that the Bayesian posterior distribution $p(\theta|X_1, ..., X_n)$ of the parameters $\theta$ given the observations $X_i$ and any reasonable Bayesian prior  approaches a Gaussian around the maximum likelihood estimate $\thetamle$: 
\begin{align}
    \|p(\theta|X_1, .... X_n) - N(\thetamlen, \frac{1}{n}\Fisher{\thetatrue}^{-1}) \|_{TV} \to 0
\end{align}
In particular it says that for large enough $n$ this holds for any reasonable Bayesian prior (that is continuous and non-zero in a neighbourhood of $\thetatrue$). Convergence is with respect to the total variation norm (i.e. the worst case event) and in probability (i.e. the chance of the TV norm exceeding any fixed threshold $\epsilon>0$ converges to zero as the number of samples $X_i\sim \modelf{\thetatrue}$ increases).

The theorem makes further assumptions about parametric models that are standard in frequentist statistics to ensure a convergence rate of the maximum likelihood estimator. Most notable are: $\modelf{\theta}(X)$ should be nonzero at all $\theta$ and $X$, $\modelf{\theta}$ twice continuously differentiable wrt. $\theta$ at all $\theta$ and $X$, the Fisher information at $\thetatrue$ should be non-singular. Additionally it assumes a uniformly consistent estimator \citep[see][for full treatment]{nickl2012:statistical} which exists for instance if among other conditions $\theta\neq\theta'$ implies $\modelf{\theta}\neq \modelf{\theta'}$~\citep{vandervaart1998:asymptotic}. Naturally not all of the above conditions can be satisfied with neural networks. In particular the weights of a neural network $\theta$ can be permuted while preserving the same function hence violating the final condition -- a property that may be amplified with ReLU activations that map an entire input range to zero. Empirically we observed better results with Leaky-ReLU activations.

\paragraph{Laplace Approximation}
The Laplace approximation is used in Bayesian statistics to approximate the posterior with a Gaussian distribution centered at the maximum a posteriori estimate $\thetamapn \defeq \argmax_\theta \prod_{i=1}^n \modelf{\theta}(X_i) p(\theta)$. It uses a second order Taylor expansion of the negative log-posterior for which it requires the Hessian $\Hessian{\theta}\defeq -\gradtheta^2 \log p(\theta|\Data)$.
$$ p(\theta|\Data) \approx \normal{\thetamapn, \Hessian{\thetamapn}^{-1}} $$
Unfortunately the Hessian is difficult to estimate and may not be positive definite hence it is frequently approximated by one of the following: the generalized Gauss-Newton matrix, the empirical Fisher or a Fisher information matrix estimate at $\thetamap$ -- see~\citet{Martens:2014,Botev2017:GaussNewton,Murphy2023:pbml} for details. 
The Laplace approximation as a tool for integration dates back to~\citet{laplace1774:integral}, has been applied to neural networks by~\citet{MacKay:92b}, and can be applied to large neural networks using K-FAC~\citep{ritter2018:laplace,daxberger2021:laplace}. When originally proposed by Laplace the function (in our case the posterior) was assumed to have a single maximum so that the Gaussian shape can cover it well.

\section{Kronecker Factorizing the Fisher information}
In the context of a second order optimization \citet{Martens:2014} proposes to estimate the Fisher information matrix of neural networks using a block diagonal structure where each block corresponds to a layer in the neural network. Furthermore each block is represented by the Kronecker product of two matrices $A \otimes G$ that are of similar size as the weight matrix $W$ of that layer ($a\times a$ and $g\times g$ for a $g\times a$ sized $W$). As a consequence the memory and computational complexity of Fisher estimation is similar to the regular gradient computation. The proposed factorization assumes that the gradient per layer is statistically independent of the gradients from all other layers. This permits efficient updates of $A$ and $G$ during the forward and backward pass of the backpropagation algorithm for neural networks. The K-FAC approach supports linear, convolutional and recurrent layers by making further assumptions about the statistical independence of the layers input activations and the gradients of the layers outputs see~\citet{Martens:2014,Grosse2016:kfacconvolution,martens:2018kfacrecurrent} for details.

This factorization does not only permit efficient estimation, but also efficient inversion and sampling. 
\paragraph{Useful Kronecker Identities}
Let us recall the the following identities \citep[see][for an overview]{Schaecke:2013kronecker}:
Matrix-vector products can be written using the operator $\mattovec$ that reshapes matrices into vectors and $\vectomat$ that shapes vectors back into matrices:
$$ (A \otimes G) x = \mattovec(G \vectomat(x) \trans{A}) $$
Inversion can be computed on each factor:
$$ (A \otimes G)^{-1} = A^{-1} \otimes G^{-1} $$
Eigendecompositions can be computed on each factor, where $\Lambda$ is a diagonal matrix containing the eigenvalues and $E$ contains the eigenvectors:
\begin{align*}
    (E_A \Lambda_A E_A^{-1}) &\otimes (E_G \Lambda_G E_G^{-1}) = \\
        &(E_A \otimes E_G) (\Lambda_A \otimes \Lambda_G) (E_A \otimes E_G)^{-1}
\end{align*}
where $\Lambda_A \otimes \Lambda_G$ is a diagonal matrix containing the eigenvalues of the Kronecker product $A \otimes G$.
Now let us define a few helpful quantities: For any eigenvalue matrix $\Lambda$ let $\Lambda^{-1/2}$ be the diagonal matrix with the inverse of the square root of $\Lambda$s eigenvalues: $(\Lambda^{-1/2})_{i,i} = 1/\sqrt{\Lambda_{i,i}}$. Finally define $B_A \defeq E_A \Lambda^{-1/2}_A$ and $B_G \defeq E_G \Lambda^{-1/2}_G$.

\paragraph{Efficient Sampling}
Given the identities from above, we can now sample from $\normal{\mu, (A \otimes G)^{-1}}$ as follows:
\begin{align*}
    z &\sim \normal{0, I_{ag \times ag}} \\
    \theta' &= \mu + (B_A \otimes B_G) z 
    = \mu + \mattovec(B_G \vectomat(z) \trans{B_A})
\end{align*}
Note that this involves only products of matrices with at most $\max(a, g)$ rows and columns. It also requires eigenvalue decomposition of $A$ and $G$ which can be amortized as in K-FAC over multiple iterations.

\section{Distributional RL}
Distributional RL \cite{Bellemare2017distributional} learns a distribution over returns $\mcreturn$. Originally referred to as the `value distribution'  \citet{rowland2018:categorical} argues that `return distribution function' is a technically more correct name. The latter name emphasizes the difference between returns and values: Recall from the introduction that learning a distribution over returns captures aleatoric uncertainty while a distribution over values captures epistemic uncertainty. By learning a distribution over returns distributional RL captures aleatoric uncertainty.

\citet{Bellemare2017distributional} introduce a distributional Bellman operator and prove its convergence for policy evaluation with respect to the Wasserstein metric. As the Wassertein metric can not be estimated empirically from sampled transitions a sample Bellman update is proposed that minimizes the KL distance between the predicted return distribution $\modelf{\theta}(\mcreturn|s,a)$ and the target that estimates $p(\mcreturn|s, a)$.
See \citet{Bellemare2017distributional} for details and note that they use different notation: The learnt distribution over returns that we call $f_\theta(\mcreturn|s,a)$ corresponds to the random value $Z_{\theta}(s, a) $ (a random value whose distribution is parameterized by $\theta$ e.g. through a neural network).
In particular they consider a categorical representation for $f_\theta(\mcreturn|s,a)$. By mapping the returns into 51 intervals the model resembles multi-class classification. While the categories result in a discretization error the empirical performance of the corresponding `C51' agent was state-of-the-art at the time. 

Recall that our example agent does not use the categorical return model from C51 and hence does not reap its empirical benefits that~\citet{Bellemare2017distributional} attribute to a combination of representation learning, reduced state-aliasing and better optimization properties of the categorical loss. Instead we use the Gaussian model, whose loss resembles the regular Q-Learning loss (see 
Section~\ref{sec:optional_variance_reduction}). We leave the combination of both approaches to future work.

\section{Miscellaneous}
\label{sec:misc_deep_sea_threshold}
\paragraph{Deep Sea Evaluation Metric}
Note that the Bsuite developers recently fixed a typo in the Github implementation. This bugfix makes the Deep Sea task easier by reducing the threshold from $20\%$ to only $10\%$ required optimal episodes. Note that this change impacts comparability of papers before and after the change. All experiments and figures in this paper use the more difficult setting that requires $20\%$ successes among all episodes. The presented results are hence directly comparable to all previous publications that used Bsuite but not necessarily to Bsuite experiments conducted after October 5th, 2022.

\end{document}